\documentclass[a4paper,fleqn]{cas-sc}
\usepackage[numbers]{natbib}
\usepackage{xcolor}
\usepackage[utf8]{inputenc}
\definecolor{linkblue}{RGB}{0,102,204}

\usepackage{hyperref}
\makeatletter
\def\orcid#1{}

\makeatother

\makeatletter
\def\@ead@symbol{}
\makeatother
\def\tsc#1{\csdef{#1}{\textsc{\lowercase{#1}}\xspace}}
\tsc{WGM}
\tsc{QE}
\tsc{EP}
\tsc{PMS}
\tsc{BEC}
\tsc{DE}

\begin{document}
    \let\WriteBookmarks\relax
    \def\floatpagepagefraction{1}
    \def\textpagefraction{.001}
    \shorttitle{SOLen}
    
    \shortauthors{Cafiso et~al.}
    
    \title [mode = title]{3D-printed Soft Optical sensor with a Lens (SOLen) for light guidance in mechanosensing}                      
    
    

    %
    \author[1]{\textcolor{black}{Diana Cafiso}}
    \fnmark[1]
    \cormark[1]
    \ead{diana.cafiso@iit.it}
    \credit{Writing – review \& editing, Writing – original draft, Visualization, Validation, Methodology, Investigation, Formal analysis, Data curation, Conceptualization, Material development and characterization}
    
    \affiliation[1]{organization={Istituto Italiano di Tecnologia, Soft BioRobotics and Perception},
        addressline={Morego 30}, 
        city={Genoa},
        postcode={16163}, 
        country={Italy}}
    
    \author[1,2]{\textcolor{black}{Petr Trunin}}
    \fnmark[1]
    \credit{Writing – review \& editing, Writing – original draft, Visualization, Validation, Methodology, Investigation, Formal analysis, Data curation, Conceptualization, Testing}
    
    \author[1]{\textcolor{black}{Carolina Gay}}%
    \credit{Writing – review \& editing, Investigation, Simulation, Hardware, Testing}
    
    \affiliation[2]{organization={The Open University Affiliated Research Centre at Istituto Italiano di Tecnologia (ARC@IIT)},
         addressline={Istituto Italiano di Tecnologia}, 
        city={Genova},
       country={Italy}}
    
    \author[1]{\textcolor{black}{Lucia Beccai}}
    \cormark[1]
    \ead[URL]{lucia.beccai@iit.it}
    \credit{Writing – review \& editing, Supervision, Resources, Project administration, Funding acquisition}
    
    \cortext[cor2]{Correspondence to: Via Morego 30, Genova 16163, Italy}
    
    \fntext[fn1]{Equally contributing authors}
    
    
    \begin{abstract}
    Additive manufacturing is enabling soft robots with increasingly complex geometries, creating a demand for sensing solutions that remain compatible with single-material, one-step fabrication. Optical soft sensors are attractive for monolithic printing, but their performance is often degraded by uncontrolled light propagation (ambient coupling, leakage, scattering), while common mitigation strategies typically require multimaterial interfaces. Here, we present an approach for 3D printed soft optical sensing (SOLen), in which a printed lens is placed in front of an emitter within a Y-shaped waveguide. The sensing mechanism relies on deformation-induced lens rotation and focal-spot translation, redistributing optical power between the two branches to generate a differential output that encodes both motion direction and amplitude. An acrylate polyurethane resin was modified with lauryl acrylate to improve compliance and optical transmittance, and single-layer optical characterization was used to derive wavelength-dependent refractive index and transmittance while minimizing DLP layer-related artifacts. The measured refractive index was used in simulations to design a lens profile for a target focal distance, which was then printed with sub-millimeter fidelity. Rotational tests demonstrated reproducible branch-selective signal switching over multiple cycles. These results establish a transferable material-to-optics workflow for soft optical sensors with lens with new functionalities for next-generation soft robots.
    \end{abstract}


\begin{keywords}
optical sensing \sep soft sensor \sep Soft robotics \sep 3D Printing \sep DLP \sep Photopolymer \sep Lens
\end{keywords}

\maketitle

\section{Introduction}
Additive manufacturing (3D printing) is increasingly used in soft robotics because it enables geometrically complex, compliant structures with integrated actuation and sensing pathways that are difficult to realize with conventional molding \cite{trunin_melegros_2026} \cite{guan_lattice_2025}. As soft robots adopt more intricate architectures and achieve richer deformation modes, sensorization must also evolve toward higher spatial resolution, greater deformability, and improved integration with the robot body\cite{hegde_sensing_2023}.

Among the emerging sensing strategies in soft robotics, optical transduction has gained attention due to comparatively simple material and process requirements. Although the mechanism is well established\cite{preti_hexsows_2025}, \cite{zhao_optoelectronically_2016}, 3D printing of soft optical sensors has just recently been introduced \cite{trunin_design_2025} \cite{tang_soft_2024}. In its basic form, an optical soft sensor can be produced by making waveguides from a transparent elastomer and then converting deformation-induced changes in light intensity\cite{lo_preti_triaxial_2024}, path length \cite{williamson_increasing_2025}, or wavelength shift\cite{bai_stretchable_2020} into an electrical signal in a photoreceiver. Because light can be routed through transparent media in complex geometries, soft optical sensors are compatible with one-step fabrication approaches in which the mechanical structure and the sensing geometry are produced simultaneously in a single material \cite{exley_monolithic_2025}.

However, the same qualities that simplify the one-step fabrication can degrade sensing performance, as light propagates in a uniformly transparent material without specific patterns. Ambient illumination may affect the sensing region and introduce noise.  Emitted light can deviate from the intended optical path, and scattering from non-ideal interfaces (surface roughness, microbubbles, or imperfect boundaries) can reduce signal-to-noise ratio and complicate calibration. Over the past decade, these issues have been mitigated primarily through reflective or opaque coatings. For instance, Lantean et al. \cite{lantean_stretchable_2022} investigated a composite elastomeric matrix with embedded reflective TiO2 particles as a coating to shield a strain sensor from external light and reduce frustrated total internal reflection. Similarly, To et al. \cite{to_soft_2018} employed a gold coating that enabled a strain-sensitive, crack-driven sensing mechanism.
Although these coating strategies have proven effective, they introduce a key limitation that restricts their scalability in soft robotics: coatings often require different materials to be interfaced. This, in turn, leads to multistep fabrication procedures that are not compatible with emerging monolithic manufacturing approaches.

A complementary route is to integrate optical elements, most notably lenses, directly within the soft sensor to shape, focus, or collimate the light field and thereby increase sensitivity and robustness. While 3D printing has been used to fabricate lenses \cite{xu_grayscale_2025}\cite{li_design_2025}, the integration of 3D-printed lenses specifically in soft optical sensors still needs investigation, in part because 3D printing of such sensors is itself a relatively recent research direction. Herein, we tailor an acrylate polyurethane resin to simultaneously improve compliance and optical transmittance, enabling the printing of the first \textbf{S}oft \textbf{O}ptical sensor with a \textbf{Len}s (SOLen). 
The proposed sensor features a Y-shape architecture consisting of one source branch connected to the emitter and two symmetric receiving branches connected to the photoreceivers. A printed lens is positioned in front of the emitter to generate a focal spot. The sensing mechanism relies on intensity modulation via focal-spot translation: bending induces a rotation of the lens, which shifts the focal spot laterally. In the undeformed configuration, the focal spot lies midway between the two receiving branches, resulting in an approximately equal light distribution and balanced signals. Upon deformation, the lens rotates and the focal spot moves accordingly, preferentially coupling light into one branch while reducing it in the other. This produces a clear, differential signal that can be correlated with both the amplitude and the direction of the motion.\\

\begin{figure*}
	\centering
		\includegraphics[width=1\columnwidth]{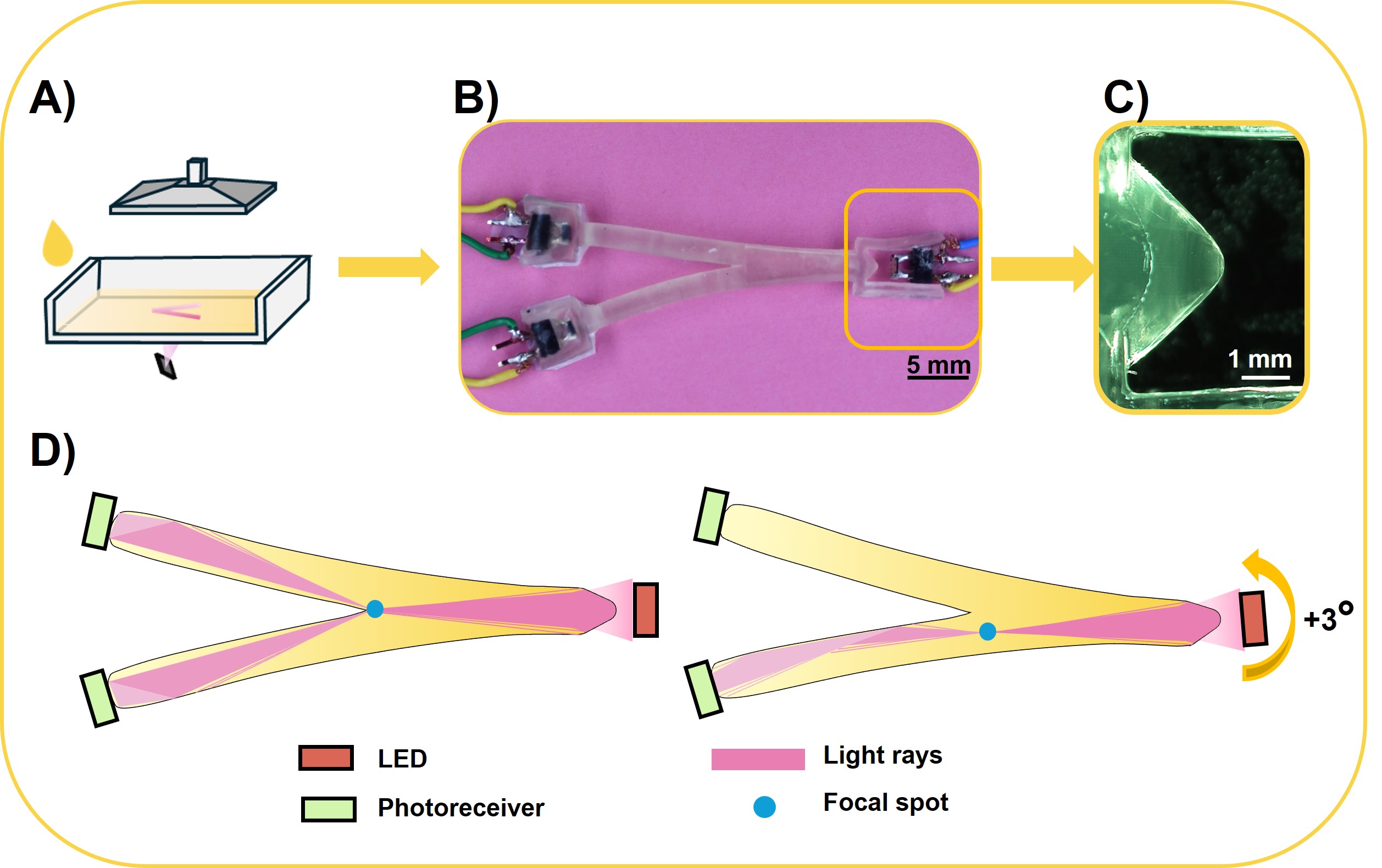}
	\caption{(A) Acrylate soft polyurethane was 3D printed via UV-based DLP to fabricate (B) SOLen sensor, with post-assembled emitter and photoreceiver. (C) Microscope image of the 3D printed lens, which is the basis of the (D) working principle of the SOLen. On the left, the sensor in the undeformed state, on the right, the sensor under rotation. The focal spot point moves with the lens, guiding the light.}
	\label{fig:figure1}
\end{figure*}

To obtain material parameters relevant for optical component design, we characterize the resin at the single-layer level, thereby minimizing refractions caused by interfacing different materials inherent to DLP printing, and measure its transmittance and refractive index. The latter is then used as input for optical simulations to define the lens profile. Leveraging the high spatial resolution of Digital Light Processing (DLP) printing, we fabricate the SOLen architecture and the integrated lens with sub-millimeter fidelity (Figure~\ref{fig:figure1}A–C). The sensing concept (Figure~\ref{fig:figure1}D, and detailed in the Results) exploits the lens rotation induced by deformation: as the lens rotates, the focal spot translates, preferentially steering light toward one receiving branch and producing a distinct signal. Overall, this work introduces the methodology (material testing, lens simulation, and sensor design) for integrating lenses into soft optical waveguide sensors, providing a route to confine and guide light in complex soft robotic geometries and thereby improving sensing precision and robustness. 

\section{Results and Discussion}

\begin{figure*}
	\centering
		\includegraphics[width=1\columnwidth]{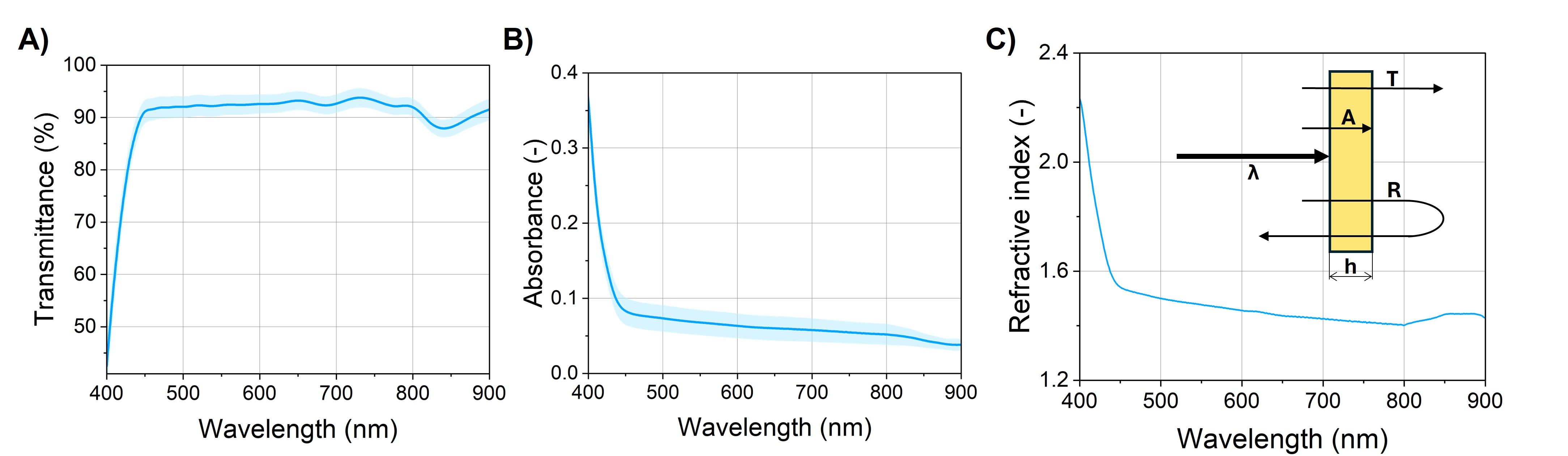}
	\caption{ UV-Vis spectra of (A) Transmittance T and (B) Absorbance A of one-layer of ma-PU (C) the analytical refractive index, measured by modeling the printed layer as a slab of thickness h. $\lambda$ is the wavelength of the incident light, and R is the reflectance. }
	\label{fig:figure2}
\end{figure*}

For optical soft sensors, the material must be transparent to specific wavelengths (according to the properties of LED and receivers) while maintaining low stiffness. Here, an acrylate polyurethane (a-PU) is modified by adding 25 wt\% lauryl acrylate (LA). Adding monofunctional monomers, such as LA, is an effective strategy to modify the mechanical properties of photocurable materials \cite{asmussen_influence_2001} \cite{cafiso_dlpprintable_2024}, leading to a decrease in crosslink density and, thus, increasing both compliance and transparency of the pristine a-PU \cite{maurya_synthesis_2017}. Mechanical and optical characterization (Figure S1A,B) confirm this trend, showing that the modified acrylate PU (ma-PU) is more deformable and more transparent. Among DLP printing parameters, the exposure time defines how long each layer is irradiated at a fixed light intensity. For a given resin composition and irradiance, exposure time governs both the photopolymerization rate and the penetration (cure) depth. At first, we determine the working curve (Figure S1C) by progressively increasing exposure time and measuring the thickness of the cured samples. The resin exhibited the typical behavior of acrylate photopolymers, namely, high reactivity and rapid curing \cite{staffova_3d_2022}. A curing energy of approximately  40 $ \mathrm{mJ/cm^{2}} $(corresponding to an exposure time of 3 s) yields a cure depth >200$\mu$  m, which is compatible with fast printing at thin layer thicknesses (e.g., 25 $\mu$m ). At lower energies, curing is still observed, but the small thickness of the specimens, combined with the low modulus of the material, make thickness measurements unreliable. Nevertheless, the working-curve trend indicates that the penetration depth remains larger than the layer thickness within the relevant printing window, which is required for robust interlayer adhesion \cite{wu_tilting_2017}. Consistent with this, preliminary prints confirm that parts can be fabricated with 25–100$\mu$m layers using exposure times below 7 s.

These results inform the selection of exposure conditions for optical characterization, where the goal is to probe changes in optical performance without introducing significant overcuring. Therefore, exposure times of 2, 4, and 6 seconds are selected. Because the lens function depends on the refractive-index contrast at the air-polymer interface, reliable values of transmittance and refractive index are essential for lens design. However, as shown in our previous work \cite{trunin_design_2025}, the layered microstructure in DLP prints can introduce additional refraction and scattering, complicating the interpretation of bulk optical measurements. To minimize these layer-induced artifacts, we characterize single-layer specimens as an approximation of a layer-less material response. UV-Vis measurements on single layers cured for 2, 4, and 6 s show a clear dependence of optical properties on exposure dose \cite{tu_additive_2025}.  In particular, as shown in Figure S1D, lower exposure produces higher transparency, consistent with a reduced crosslink density. Based on these results, we select 2 s as the exposure time for subsequent prints and prepare single-layer samples (fabricated as shown in Figure S2A-D) to measure the refractive index n, a key input for lens simulations, from the other optical properties. Overall, the material exhibits high transmittance T (>85\%, Figure~\ref{fig:figure2}A) and low absorbance A between 450 and 900 nm (Figure~\ref{fig:figure2}B), providing flexibility in selecting the LED emission wavelength. 

The refractive index is calculated by modeling the sample as a parallel-face partially absorbing slab \cite{nichelatti_complex_2002}:

\[
n(R,T)=\frac{1+R_F(R,T)}{1-R_F(R,T)}
+\left\{
\frac{4R_F(R,T)}{\left[1-R_F(R,T)\right]^2}
-\left(\frac{\lambda}{4\pi h}\right)^2
\ln^2\!\left[\frac{R_F(R,T)\,T}{\,R-R_F(R,T)\,}\right]
\right\}^{1/2}.
\]

where h is the geometrical thickness, $\lambda$ is the wavelength and $R_f$ is the reflectance coefficient at the interface (Figure S2E). 
As reported in Figure~\ref{fig:figure2}C, n is approximately 1.50, consistent with other acrylate-based polymers \cite{tang_high-refractive_2020} \cite{reynoso_refractive_2021}, and exhibits an expected wavelength dependence that yields higher values below 450 nm. To demonstrate the versatility of the approach (i.e., flexibility in selecting the light source), we select three wavelengths and use the corresponding refractive indices to simulate sensors with different lens profiles. The measured refractive index is used to inform lens design. The overall sensor layout is fixed as a waveguide with two branches (Figure~\ref{fig:figure1}D).

Finite element simulations are performed in COMSOL Multiphysics® (COMSOL Inc., Sweden). To reduce computational cost, a 2D model is used, coupling the Ray Optics module through the Multiphysics interface.
The lens is designed to produce a focal spot inside the material, located 20 mm from the lens, with the emitter placed 1 mm from the lens. A Cartesian oval is used to define the lens profile \cite{sutter_ideal_2019}:

\begin{equation}
n_1 \sqrt{(z + s)^2 + x^2}
+ n_2 \sqrt{x^2 + (s' - z)^2}
- (n_1 s + n_2 s') = 0
\end{equation}

\begin{figure*}
	\centering
		\includegraphics[width=1\columnwidth]{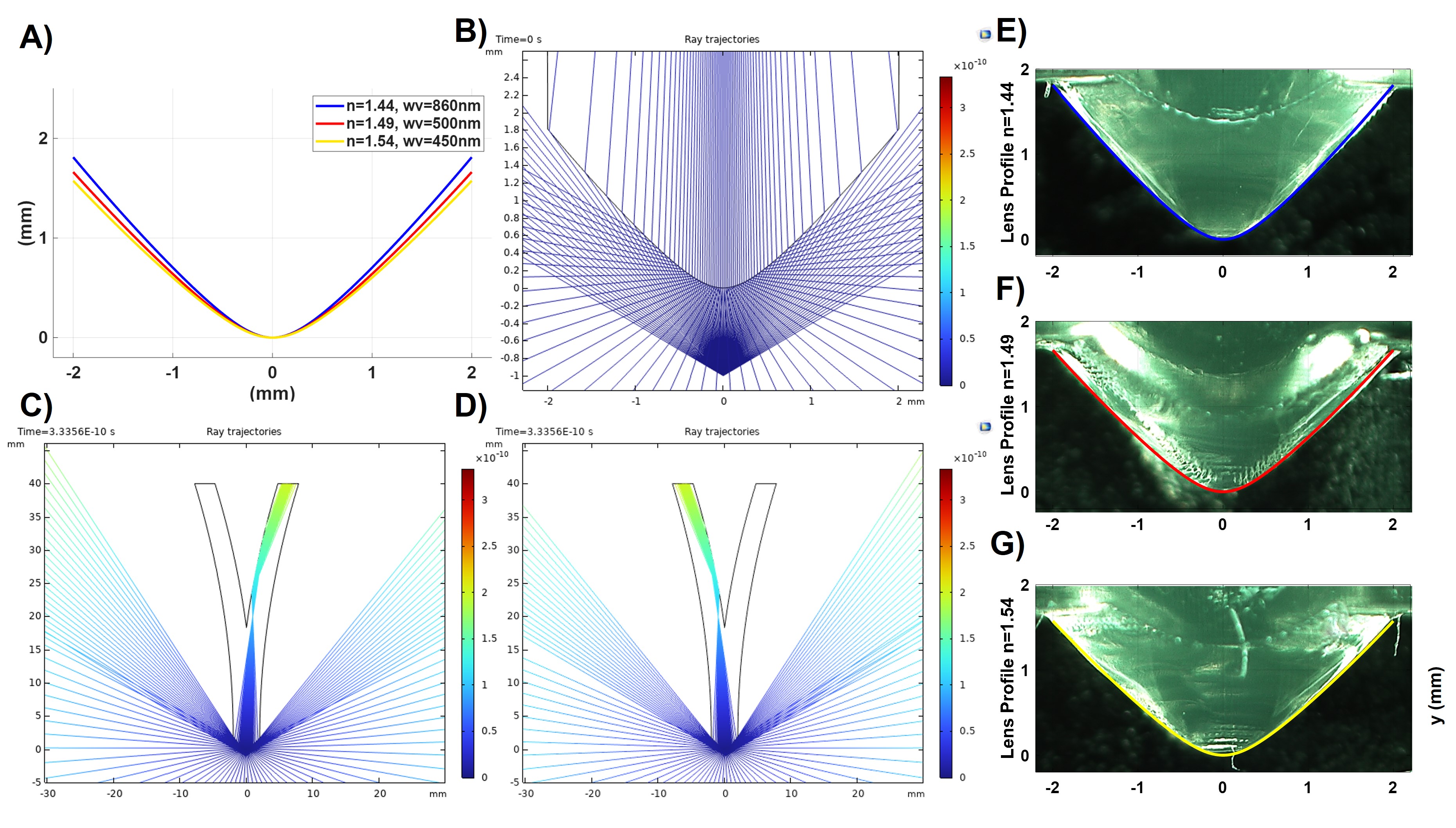}
	\caption{(A). Graph three different lens profiles obtained with different RI parameters in the Cartesian oval's equation. (B). Zoom on the lens (RI=1.44) during the simulation. (C). Simulation after the lens rotation of $-3^\circ$. (D). Simulation after the lens rotation of $+3^\circ$. (E-G) Printed lens microscope images compared with the Cartesian ovals for the different refractive index (RI in the figure).
	}
	\label{fig:figure3}
\end{figure*}

where $n_1$ is the refractive index of air, $n_2$ is the refractive index of the lens material, $s$ is the distance between the lens and the emitter, and $s^{\prime}$ is the focal distance. Here, $x$ represents the lens width (ranging from $-2$ to $2$ mm), and $z$ is the resulting lens profile.

The lens is integrated into a Y-shaped sensor geometry (Figure~\ref {fig:figure1}D). The stem is 16.2 mm long, and each arm measures 22.2 mm. To smooth the outer corners between the stem and the arms, a second-order parametric curve $(a = -0.004)$ is used. The stem forms a trapezoid with parabolic sides, with a minor base of 4 mm (lens base) and a major base of $~6.2$ mm (fork point). Each arm has a width of $~$3 mm (precise geometry is shown in Figure S3). The emitter is positioned 1 mm from the lens, and the two receivers are placed at the ends of the arms.

\begin{figure}
	\centering
		\includegraphics[width=1\columnwidth]{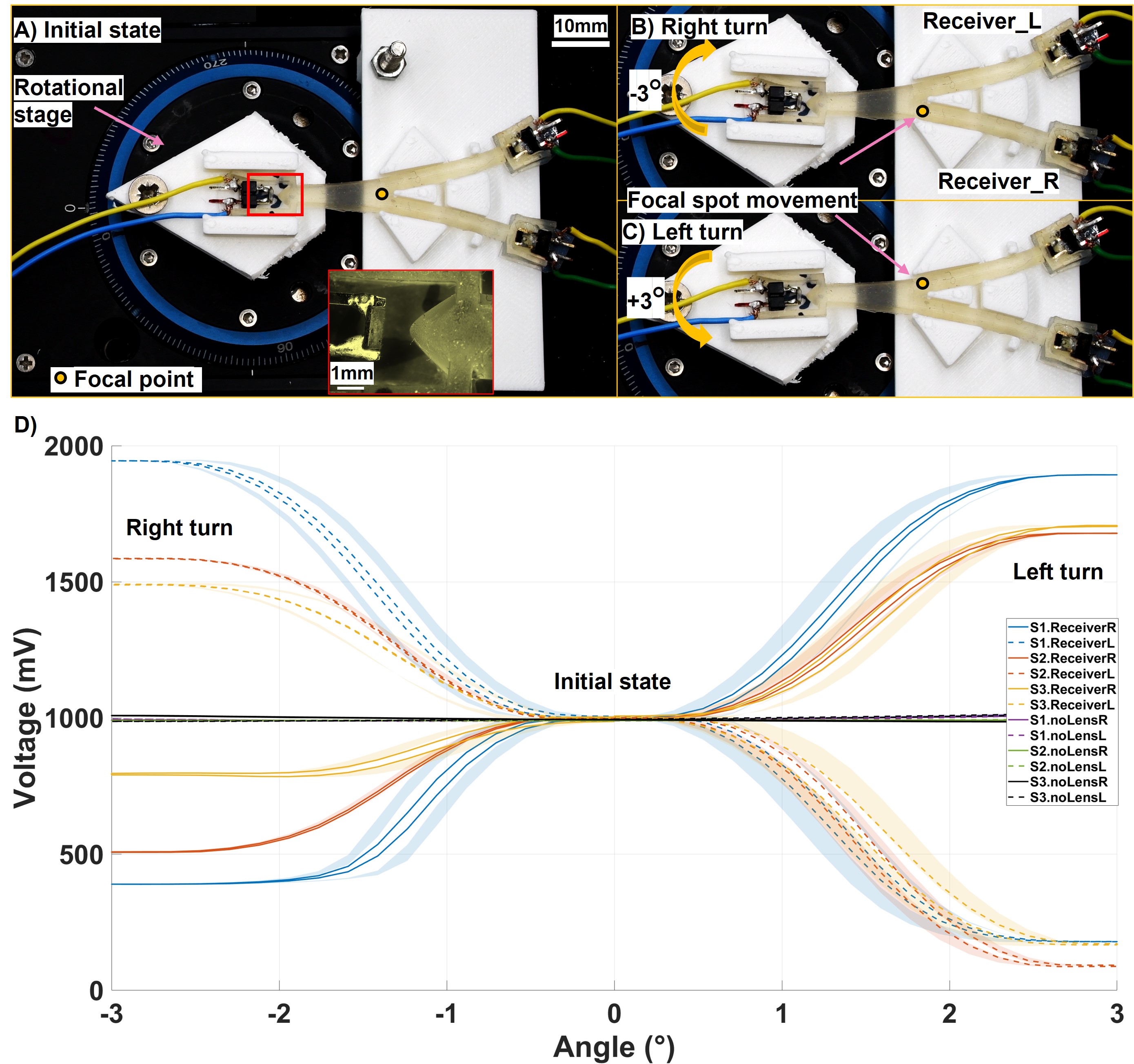}
	\caption{Figure 4. Experimental validation of lens-driven focal switching in the Y-shaped optical sensor. (A) Initial configuration of the test, where the emitter is aligned with the centerline between the two branches, placing the focal spot between ReceiverL and ReceiverR. (B) Right-rotation state (-3°), achieved by tilting the emitter/lens assembly to shift the focal spot into the right branch toward ReceiverR. (C) Left-rotation state (+3°), where the focal spot is redirected into the left branch toward ReceiverL. (D) Measured receiver signals plotted as voltage versus rotation angle for three sensors equipped with lenses over five consecutive cycles (mean curve with shaded standard deviation). A clear intensity redistribution is observed between the two receivers: during right rotation, ReceiverR shows a voltage decrease (higher intensity) while ReceiverL shows a voltage increase (lower intensity), with the opposite trend during left rotation.}
	\label{fig:figure4}
\end{figure}

To evaluate sensitivity to misalignment, both the emitter and the lens are rotated by $\pm 3^\circ$ around the center of the stem’s minor base, and the resulting focal shift is quantified. To isolate the lens's optical contribution, simulations are run with and without the lens (Figure S4). To show the versatility of the method, the sensor is simulated using three refractive indices (1.44, 1.49 and 1.54, corresponding to 860, 500 and 450 nm), resulting in modified lens profiles (Figure~\ref{fig:figure3}A). In the simulation, sensors equipped with a lens show a clear separation between the responses under the two rotation conditions (Figure~\ref{fig:figure3}B-D), whereas the difference without a lens is negligible (Figure S4G-I).

After validating the overall design, the three different lens profiles are printed to assess the fidelity of the dimensions in the fabrication. The agreement between the microscopy images and the calculated profiles is shown in Figure~\ref{fig:figure3}E-G. Despite slight deviations between the CAD and the printed parts, the three lenses exhibit clearly distinguishable surface profiles, consistent with the measured printing resolution of the resin, studied in Figure S5A-D. Specifically, the minimum resolvable size in x-y is 610 $\mu$m and 50 $\mu$m for positive and negative features, respectively. Adjacent structures are separated by 500 $\mu$m without overcuring, the resolved step in z was  25 $\mu$m, consistent with the layer thickness.  
Finally, the sensing response of printed structures is tested. For these experiments, the 860 nm emitter is used; therefore, the lens profile is based on the refractive index (1.44) at this wavelength. 

The experimental setup consists of a rotational stage with the emitter positioned at the center, enabling the experiment to closely replicate the simulation conditions. In the initial configuration, the emitter is aligned with the centerline between the two branches, so the focal spot lies between them (Figure~\ref{fig:figure4}A). During the test, the rotational stage tilts the emitter by $\pm 3^\circ$ to one side at an angular speed of 15 deg/s, holds this position for 0.8 s, and then returns to the initial alignment. Then it rotates to the opposite side, shifting the focal spot toward the other branch (Figure~\ref{fig:figure4}B,C).

The protocol is repeated for six cycles. The last five cycles are used to generate voltage-versus-rotation-angle plots. The raw voltage-versus-time traces are reported in the Supporting Information (Figure S6). Data are cropped to exclude the holding (“waiting”) phases and plotted (Figure~\ref{fig:figure4}D). Figure~\ref{fig:figure4}D shows a clear shift of the focal spot from one branch (ReceiverR) to the other (ReceiverL) across three different samples over five cycles (mean curve with shaded standard deviation). For Sensor 1 (blue curve), the transition from the initial state to the right-rotation state results in a decrease in voltage at S1.ReceiverR (corresponding to an increase in intensity) and a voltage increase at S1.ReceiverL (corresponding to a decrease in intensity). The same behavior is observed for the other two sensors with lenses, confirming the proposed working principle.

The same procedure is repeated using a sensor without a lens. As expected from both the simulations and basic optical considerations, the response without a lens is negligible, confirming that the lens is responsible for the observed signal separation.

\section{Conclusion}

In this work, we introduce a lens-based strategy to control light propagation in 3D-printed soft optical sensors. By integrating a printed lens in front of the emitter within a Y-shaped waveguide, the optical field is actively shaped: deformation-induced lens rotation produces focal-spot translation, which redistributes optical power between the two receiving branches and generates a clear differential output. This provides a complementary route to improve signal separation and direction sensitivity without relying on coatings or claddings.

Beyond the specific SOLen architecture demonstrated here, the proposed methodology is broadly adoptable across photopolymer-based soft materials. The approach combines single-layer optical characterization to obtain transmittance and refractive index values, with simulation-driven lens profiling (Cartesian-oval design) to define the required optical element for a target focal distance and emitter placement. As a result, the lens can be redesigned and reprinted for different printable polymers, exposure conditions, and wavelengths, offering a transferable pipeline for optical element integration in additively manufactured soft sensors.

This lens-integration concept complements existing soft optical waveguide sensing paradigms. It can function as a strategy when multimaterial fabrication is undesirable or impractical, and it can also be used as an additional enhancement alongside core/cladding or coatings by shaping and steering the emitted light field, improving coupling efficiency and differential readout.

\begin{figure*}
	\centering
		\includegraphics[width=1\columnwidth]{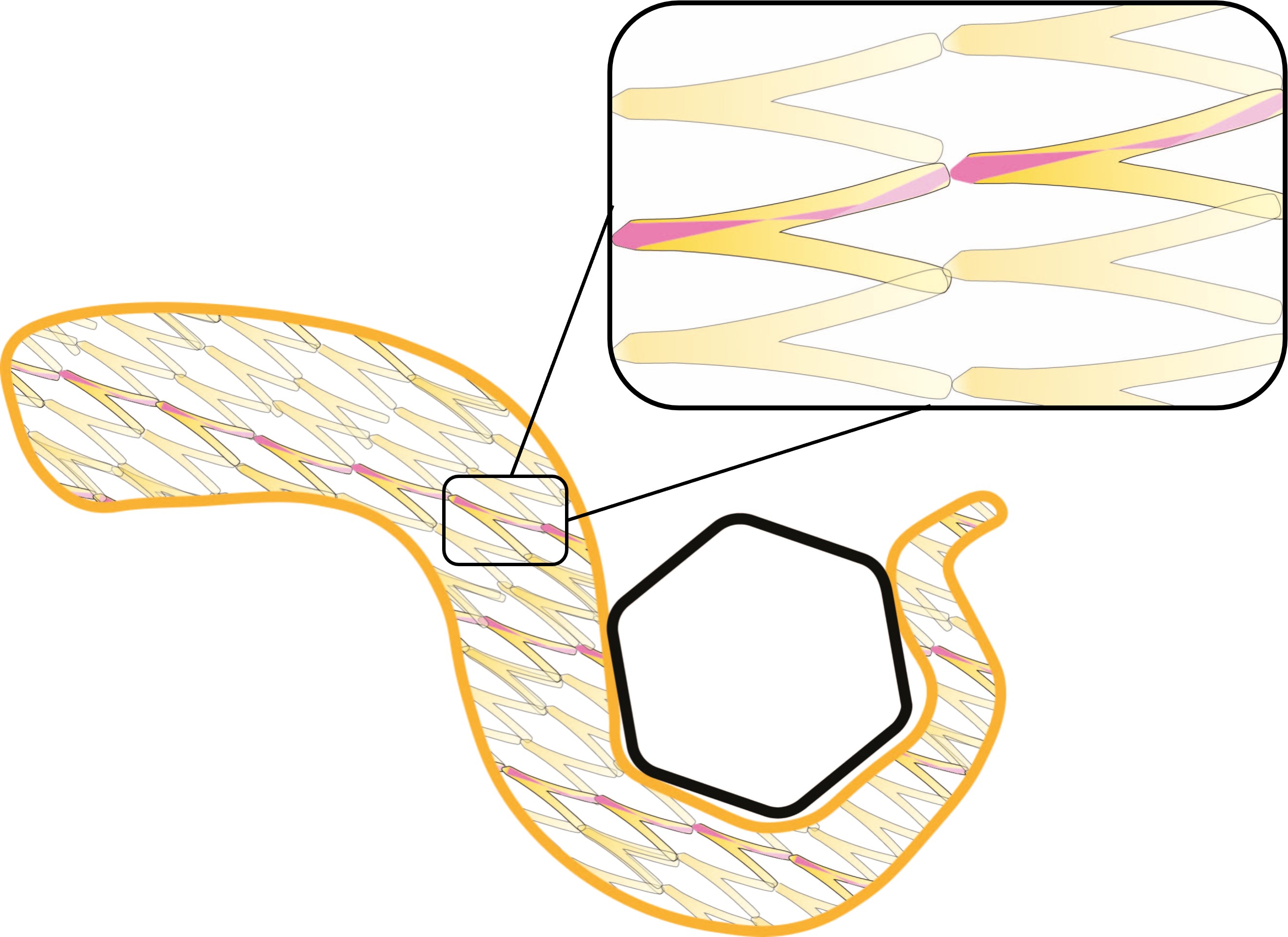}
	\caption{ Conceptual example of SOLen integration in a monolithic soft robot architecture. Selected structural elements act as waveguides, while printed lenses focus and steer light toward predefined directions to enable distributed sensing.}
	\label{fig:figure5}
\end{figure*}

Critically, lens-enabled optical control is particularly suited to emerging monolithic soft robotics systems, where single-material fabrication is a key enabler of robustness and scalable manufacturing. As soft robots continue to increase in geometric and functional complexity (\cite{trunin_melegros_2026}\cite{guan_lattice_2025}), guiding and confining light within a single transparent material provides a mechanism to expand sensing functionality without adding multimaterial interfaces or additional fabrication steps. Figure~\ref{fig:figure5} shows a representative monolithic concept in which the SOLen sensing elements form the geometry of the robot: selected structural members also function as optical waveguides, and integrated lenses steer and concentrate light toward designated receiving paths. Building on the SOLen approach, optical design driven by material characterization and simulation enables programmable internal light routing, i.e., tailoring where light travels and how it is distributed within a single transparent body to generate spatially distinct sensing responses. In this sense, the ability to architect the internal light field through printed optical elements supports a trajectory toward higher integration and richer sensorimotor coupling, which is a necessary step for the progression of soft robots toward embodied intelligence.

\section{Materials and Methods}
\subsection{Materials}
The developing acrylate polyurethanes RA-139 (a-PU) were provided by Photocentric (Peterborough, UK). Lauryl acrylate (LA) was purchased from Sigma-Aldrich (St.Louis, Missouri, US).
\subsection{Material's design and printing}
The modification of the resin was carried out by adding 25wt\% of  LA. The components were mixed under magnetic stirring for 10 minutes. The modified formulation is hereinafter referred as ma-PU. 
The  resins were 3D-printed through an Asiga DLP 3D printer (Asiga, Alexandria, AUS) with a light source of 385 nm.
The working curves of RA and RA* were measured by the polymerization of a thin layer using the UV-projector of the printer. A drop of formulation was placed on the bottom of the vat and subjected to UV-exposure at {13}$ \mathrm{mW/cm^{2}} $ for different curing times. The projected image was a circle with a diameter of 3 mm. The depth of curing  Dc (i.e, thickness of the cured disks) was measured by means of a digital microscope (Hirox, Tokyo, JP). The critical energy (Ec) and the penetration depth (Dp) were derived from the Jacobs equation \cite{rudenko_light_2024}:

\[
D_c= D_p \ln{\frac{E}{E_c}}
\]

where E is the energy of the light beam (mJ/cm2).
The results were used to set the printing parameters (exposure time, layer thickness). 

\subsection{Optical and mechanical characterization}
The mechanical compression curves of a-PU and ma-PU were collected through compression tests (ZwickRoell, Ulm, DE) on cubic samples (1 cm x 1 cm x 1 cm). The optical transmittance (T) of a-PU and ma-PU was measured via a UV-Vis-NIR spectrophotometer Cary 5000 (Agilent, Santa Clara, US), in the range of 900-400 nm. 
One-layer samples were fabricated as follows. First, a base of RA* (0.5 mm) was printed. Then, a film of Polyethylene terephthalate (PET) sheet with a thickness of 0.1 mm was cured on the base, without removing this last from the building platform of the printer. After calibrating the printer for the new thickness, one layer of ma-PU was printed. After printing, the PET sheet (with the ma-PU one layer on top) could be easily peeled off the ma-PU base. The resulting sample consisted of PET sheet and ma-PU (one layer). During the optical characterization, the PET was used to collect the baseline and retrieve the optical transmittance and absorbance (A) of one layer of ma-PU.
The Reflectance was measured from UV-Vis spectra as:
\[
R= 1- T - A 
\]
T and R were used to calculate the refractive index n (Figure~\ref{fig:figure2}C) as follows \cite{nichelatti_complex_2002}:
\[
n(R,T)=\frac{1+R_F(R,T)}{1-R_F(R,T)}
+\left\{
\frac{4R_F(R,T)}{\left[1-R_F(R,T)\right]^2}
-\left(\frac{\lambda}{4\pi h}\right)^2
\ln^2\!\left[\frac{R_F(R,T)\,T}{\,R-R_F(R,T)\,}\right]
\right\}^{1/2}.
\]

In which the sample is regarded as a partially absorbing slab with a plane-parallel face. In the equation, h is the geometrical thickness, $\lambda$ is the wavelength and $R_f$ is the reflectance coefficient at the interface, measured as:
\[
R_F=\left(2+T^{2}-(1-R)^{2}
-\left\{\left[2+T^{2}-(1-R)^{2}\right]^{2}-4R(2-R)\right\}^{1/2}\right)
\left(2(2-R)\right)^{-1}.
\]

The printing fidelity of the resin was quantified printing and measuring dedicated parts to evaluate the minimum resolvable positive (raised) and negative (recessed) features in the XY plane, as well as the minimum reproducible step in Z. Positive-feature fidelity was evaluated using arrays of raised cylindrical features on a base of 0.8 mm, with systematically varied lateral size and inter-feature spacing (Groups A and B, described below), and a stepped-height structure to assess Z discretization (Group C, described below). Negative-feature fidelity was evaluated using a perforated slab containing circular through-holes with decreasing diameters. All printed parts were inspected by optical microscopy and used to extract: (i) the minimum resolvable XY feature size for raised elements, (ii) the minimum XY separation required to avoid feature merging, (iii) the minimum resolvable diameter variation for recessed features, and (iv) the minimum reproducible Z step. The geometries of the different parts are detailed as follows:

\begin{itemize}
        \item Group A: All cylinders have a fixed height of 3.0 mm, while the nominal diameters are: 1.00, 0.80, 0.60, 0.50, 0.40, 0.30, 0.25, 0.20, 0.18, 0.16, 0.14, 0.12, 0.10, 0.08, and 0.06 mm.
        \item Group B: cylinders with fixed nominal diameter 0.30 mm and height 3.0 mm, arranged with systematically varied edge-to-edge gap values. The nominal gaps are: 0.50, 0.30, 0.20, 0.15, 0.12, 0.10, 0.08, 0.06, 0.05, 0.04, 0.03, and 0.02 mm.
        \item Groups C: cylinders of 0.50 mm nominal diameter with heights 0.1 mm - 1 mm (0.1 mm steps), heights 0.2 mm - 0.5 mm (0.5 mm steps), heights 0.2 mm - 0.5 mm (0.25 steps)
        \item Negative-features: a rectangular slab (16 × 3 × 1.00 mm) containing eight circular through-holes arranged in a single row with nominal hole diameters span 1.40–1.65 mm (steps of 0.1 mm)
\end{itemize}

To provide a visible reference under the microscope, an additional artifact was printed. The model comprises a base slab ( 10 × 3.2 × 0.80 mm) supporting three pillars (each  3.00 × 3.00 mm in planform) with heights of 1.000 mm, 1.025 mm, and 0.025 mm, respectively. This design facilitates consistent focusing and enables visual confirmation of small Z offsets ( 25 µm) under the same imaging conditions used for the resolution parts.

\subsection{Sensor's design and simulation}

Discrete points representing the lens profile were generated in Python and imported into a 2D COMSOL Multiphysics® model (COMSOL Inc., Sweden) using the Interpolation Curve feature to reconstruct the continuous profile. The sensor mesh was automatically generated by the software based on a normal distribution.

In the simulation setup, the emitter was positioned 1 mm from the lens, and the two receivers were placed at the ends of the arms. Within the Ray Optics module, the emitter was modeled as a point source emitting 100 rays in a conical distribution with an aperture angle of $120^\circ$, according to the emitter specifications. The receivers were defined as segments at the extremities of the arms. The wavelength of the light was initially set to 860 nm and adjusted according to the corresponding refractive index.

Lens rotation was simulated by rotating both the lens and emitter within the Geometry module, allowing evaluation of their relative orientation while maintaining rapid computation. The number of detected rays was recorded for both receivers during the simulations.

\subsection{Testing}
All the tests were conducted using a custom platform, as reported in Figure~\ref{fig:figure4}A. A micrometric servo-controlled rotation stage M-060.DG (Physik Instrumente, Karlsruhe, Germany) with 3D printed additional parts and screws were used for the tests. The responses of the two photoreceivers were captured using a DAQ system (USB-6218, National Instruments, Austin, TX, USA). All the data was analyzed in Matlab(The MathWorks, Inc).
The MPU tests were conducted using an M-111.1DG micrometric servo-controlled linear stage (Physik Instrumente, Karlsruhe, Germany) with additional 3D printed parts.
\subsection{Custom experimental equipment}
Custom experimental fixtures for the two-point rotation test were printed by means of a Fused Filament Fabrication (FFF) 3D printer (Ultimaker S3, Ultimaker, Utrecht, NE) by extruding ABS.

\subsection{Fabrication and characterization of the sensors}
Sensors were printed with a layer thickness of 25 $\mu$m, using six burn-in layers (exposure time of 3 seconds) and an exposure time of 2 seconds. The light intensity was set as 13 mW/cm2. After printing, the part was delicately washed with isopropanol to remove the uncured resin. To precisely embed the electronic components (photoemitter PE and photoreceivers PRs), supports were cut from a PET sheet (0.1 mm thick) and glued to the extremities of the sensors (see SI XX figure). Some resin was cured to the PRs to ensure their stable embedment. Nothing was poured on the emitter to alter the lens profile and maintain the "air" medium between the light source and the lens. Wires were welded to each PR and to the PE.  The lenses were charcaterized by microscope to assess the printing fidelity (Hirox, Tokyo, JP). 

\subsection{Sensing setup}
The photoemitters used in the devices (infrared LEDs, VSMY1850) have a peak emission at 860 nm, corresponding to the peak reception of the photoreceivers(TEMT7100X01). Signals were acquired using a custom-designed PCB. The PCB enables sequential activation of the emitters, thereby minimizing interference from ambient light. The emitters are modulated at 200 Hz, allowing the system to record the ambient light level during the “off” phase and subtract it from the measurements taken during the “on” phase. In addition to the transistor circuitry, the PCB incorporates tunable resistors (TC33X-2-202E), which allow adjustment of receiver sensitivity and emitter power to accommodate variations in sensor geometry and mechanical deformation. Data collection was done with Python 3.13, and data analysis was performed using MATLAB (The MathWorks, Inc.).

\subsection{Statistical analysis}
All sensing data was processed using MATLAB (MathWorks). Data was not smoothed or filtered. No formal hypothesis testing or between-group statistical comparisons were performed; all statistics are descriptive and intended to demonstrate repeatability and variability across samples. All data processing and statistical computations were performed in MATLAB (MathWorks).

\printcredits

\bibliographystyle{unsrtnat}

\bibliography{references}

\end{document}